\documentclass{article}
\usepackage{spconf,amsmath,graphicx}
\usepackage{amssymb,amsfonts}
\usepackage{cite}
\usepackage[hidelinks]{hyperref}
\usepackage{microtype}
\usepackage{balance}

\title{Joint Estimation of Common-Slope Decay Rates and Spatial Amplitudes using Parameterized Nonnegative Matrix Factorization}

\name{Jeremy B. Bai$\,^\star$, Filip Elvander$\,^\dagger$, Sebastian J. Schlecht$\,^\star$}
\address{$\,^\star$Multimedia Comms. \& Signal Proc., Friedrich-Alexander-Universität Erlangen-Nürnberg, Germany \thanks{JBB and SJS were supported through the joint German Academic Exchange Service (DAAD) and Research Council of Finland (RCF) Project (57763119) ``Immersive Augmented Acoustics (IAA)''.}\\
    $\,^\dagger$Dept. of Information and Communications Engineering, Aalto University, Finland}

\begin{document}
\ninept
\maketitle

\begin{abstract}
We formulate joint estimation of common-slope decay rates and amplitudes from room impulse responses (RIRs) as parameterized nonnegative matrix factorization with the Itakura--Saito divergence as the loss function (IS-NMF). Estimation at each short-time Fourier transform frequency bin produces detailed reverberation time (RT) curves directly from RIR powers with no backward integration needed. Standard space-alternating generalized expectation-maximization (SAGE) algorithm yields closed-form amplitude updates and a convex subproblem for each decay rate update.  To accelerate estimation, we introduce contribution-weighted SAGE, which emphasizes observations where each component contributes strongly to the modeled power. Experiments with synthetic data show accurate recovery of well-separated decays and faster loss reduction than standard SAGE. Application to measured coupled-room RIRs yields frequency-dependent RT curves and reveals complementary space–time contributions of the shared decay components.
\end{abstract}


\section{Introduction}

Reverberation time (RT) describes how fast sound energy decays in a room impulse response (RIR). \cite{ISO3382RoomAcoustic2009} establishes RT as a standard parameter for characterizing room acoustics, while \cite{valimakiFiftyYears2012} reviews its role in modeling and synthesizing artificial reverberation. Reliable frequency-dependent RT estimation therefore supports both room-acoustic analysis and late-reverberation synthesis. \cite{schroederNewMethodMeasuring1965} provides the basis for estimating RT from measured RIRs through backward integration. \cite{jotanalysissynthesisapproach1992} extends this principle to frequency-dependent decay analysis through the Energy Decay Relief and connects it directly to reverberation synthesis. \cite{xiangEvaluationReverberationTimes1995,karjalainenEstimationModalDecay} subsequently develop estimators that account for background noise.

\cite{kuttruffRoomAcoustics2006} shows that coupled rooms and rooms with non-uniform absorption can exhibit sound-energy decays that depart from a single exponential. \cite{xiangEvaluationDecayTimes2001,xiangBayesianCharacterizationMultipleslope2011} estimate multi-exponential decay parameters for individual RIRs through Bayesian inference. \cite{gotzNeuralNetworkMultiexponential2022} introduces DecayFitNet for efficient neural estimation of those parameters. \cite{gotzCommonSlopeModelingLate2023} subsequently uses its estimates in common-slope analysis, which assumes shared RTs across sound source or receiver positions with spatially varying amplitudes.
Pooling observations across RIRs of different positions can help identify components that are weak in individual RIRs. 
The analysis in \cite{gotzCommonSlopeModelingLate2023} clusters the individually estimated RTs using $k$-means, then fits amplitudes by constrained least squares with the common slopes fixed. These approaches operate on backward-integrated energy decay curves. \cite{baiEstimationMultiSlopeAmplitudes2025} develops likelihood-based amplitude estimation with prescribed RTs using a LinEx loss on log energy. These approaches leave the joint statistical estimation of shared decay rates and spatial amplitudes unaddressed.

We jointly estimate shared decay rates and spatial amplitudes from short-time Fourier transform (STFT) powers. We formulate this task as nonnegative matrix factorization, with temporal features parameterized by exponential decay rates and Itakura--Saito divergence as loss function (IS-NMF). Independent frequency-bin estimation uses all RIRs to estimate shared decay rates without a separate clustering stage. \cite{FevotteNonnegativematrix2009} interprets this factorization as variance estimation for superimposed complex Gaussian components. We specialize its variance model to common-slope estimation using space-alternating generalized expectation-maximization (SAGE)~\cite{FesslerSpacealternatinggeneralized1994,fevotteNonnegativeMatrixFactorization2009}. Profiling out the amplitudes yields a convex scalar subproblem for each decay rate, which can be solved with safeguarded Newton updates.

Second, we propose contribution-weighted SAGE (CW-SAGE), which accelerates decay rate estimation by emphasizing samples where a component contributes strongly. This reduces the influence of samples whose posterior component power is dominated by uncertainty and therefore contribute little to parameter updates~\cite{fevotteNonnegativeMatrixFactorization2009}.  Related work on EM for Gaussian mixture models examines slow convergence and alternative update strategies~\cite{MaAsymptoticConvergence2000,NaimConvergenceEM2012,LuckeTruncatedVariational2019}. These studies perform reweighting in different formulations, but provide heuristic motivation for our proposed method in an additive Gaussian model.

\begin{figure*}[!htbp]
 \centering
 \includegraphics[width=\textwidth]{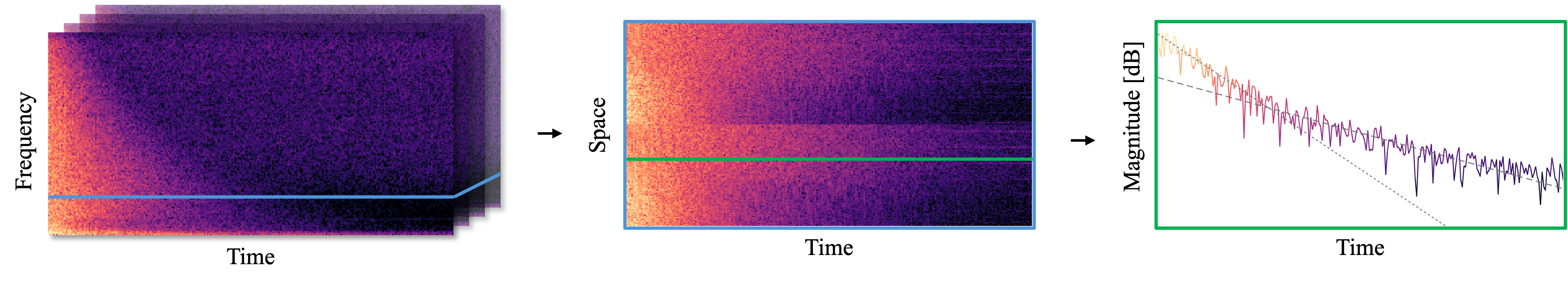}
 \caption{Parameterized NMF for common-slope estimation. A fixed-frequency slice of all the stacked RIR spectrograms (in blue frame) forms a space--time power matrix $\bf Y$. Its expected power $\bf V$ is factored into nonnegative spatial amplitudes $\bf A$ and exponential temporal features $\bf \Phi$ with decay rates shared across RIRs. The right panel shows an individual observed response (in green frame)  and the two fitted exponential decay components shown as straight lines in dB scale.}
 \label{fig:time-space-factorization}
\end{figure*}

Section~\ref{sec:method} formulates the problem and introduces the estimation algorithm. Section~\ref{sec:result} presents experiments with synthetic data and an application to measured RIRs. Section~\ref{sec:discussion} summarizes our findings and discusses limitations and future work.

\section{Methodology}
\label{sec:method}

\subsection{Parameterized NMF for common-slope modeling}

Given the STFT of a batch of RIRs from the same enclosed space with different sound source and receiver positions, let $r=1,\ldots,R$ index the spatial configurations, $f=1,\ldots,F$ frequency bins, and $n=1,\ldots,N$ time frames after a selected late-reverberation onset. For STFT coefficients $X_{rfn}$, assume
\begin{equation}
 X_{rfn}\sim\mathcal{CN}(0,V_{rfn}),\qquad
 V_{rfn}=\sum_{k=1}^{K}a_{rfk}e^{-\lambda_{fk}\tau_n},
 \label{eq:model}
\end{equation}
where $\tau_n\geq0$ is elapsed time, $a_{rfk}\geq0$ is initial power amplitude, and $\lambda_{fk}>0$ is decay rate shared across RIRs. Independence between $X_{rfn}$ is assumed across all indices. The number of components $K$ is prescribed. We treat frequency bins independently and \emph{omit subscript $f$ from this point onward}.

At one frequency, collect powers of observations $Y_{rn}=|X_{rn}|^2$ into $\mathbf Y\in\mathbb R_+^{R\times N}$ and define
\begin{equation}
 \mathbf V=\mathbf A\boldsymbol\Phi(\boldsymbol\lambda),\quad 
 [\mathbf A]_{rk}=a_{rk},\quad
 [\boldsymbol\Phi]_{kn}=e^{-\lambda_k\tau_n}.
 \label{eq:factorization}
\end{equation}
The columns of $\mathbf A$ contain spatial amplitudes of each component, and the rows of $\boldsymbol\Phi$ are their exponentially decaying temporal features (see Fig.~\ref{fig:time-space-factorization}). Specifying $K$ decay rates determines all $K\times N$ entries in $\boldsymbol\Phi$. Each exponential equals one at $\tau=0$, fixing its scale; $a_{rk}$ is the component power at that origin. The 60-dB reverberation time (RT) of decay component $k$ is
\begin{equation}
 T_{k}=\frac{6\ln 10}{\lambda_k}.
 \label{eq:t60}
\end{equation}
A stationary noise floor can be included by adding $k=0$, with $\lambda_0=0$. Only amplitudes of the noise floor will be estimated.

Writing the negative log-likelihood for the STFT powers, the complex Gaussian model gives loss function~\cite{fevotteNonnegativeMatrixFactorization2009}
\begin{equation}
 \mathcal L(\mathbf A, \boldsymbol \lambda) = \sum_{r,n} \left(\frac{Y_{rn}}{V_{rn}} + \ln V_{rn}\right). \label{eq:loss}
\end{equation}
The associated Itakura--Saito divergence is
\begin{equation}
     d_{\mathrm{IS}}(y\mid v)=\frac{y}{v}-\ln\frac{y}{v}-1. \label{eq:is}
\end{equation}
The loss in \eqref{eq:loss} equals the summed IS divergence between observed powers and modeled variance $d_{\mathrm{IS}}(\mathbf Y\mid \mathbf V)$  up to an additive data-dependent constant. The corresponding loss in log energy has the LinEx form used in our previous work~\cite{baiEstimationMultiSlopeAmplitudes2025}. We fit powers without backward integration. The STFT window sets the frequency resolution.

\subsection{SAGE algorithm for optimization}

SAGE is an algorithm within the expectation-maximization (EM) family. Instead of updating all parameters simultaneously during M-step, SAGE updates one component at a time while holding the others fixed~\cite{FesslerSpacealternatinggeneralized1994}. 

Introduce independent latent coefficients
\begin{equation}
    Z_{rnk}\sim\mathcal{CN}(0,v_{rnk}),\qquad
    v_{rnk}=a_{rk}e^{-\lambda_k\tau_n}.
    \label{eq:latent}
\end{equation}
The latent variables serve as unobserved components of the observed data $ X_{rn} = \sum_k Z_{rnk} $ from \eqref{eq:model}. At the current parameters, the component's Wiener-style posterior contribution ratio, defined as its share of the modeled variance, and its posterior power are
\begin{align}
 \rho_{rnk}&=\frac{v_{rnk}}{V_{rn}}\in [0,1],\label{eq:ratio}\\
 \hat{y}_{rnk}&=\mathbb E\!\left[|Z_{rnk}|^2\mid X_{rn}\right] =\rho_{rnk}^{2}Y_{rn}+v_{rnk}(1-\rho_{rnk}).
 \label{eq:estep}
\end{align}
Higher \(\rho_{rnk}\) reveals the space--time regions in which each fitted decay component $k$ contributes strongly to the modeled power. The second term in \eqref{eq:estep} accounts for conditional uncertainty. This corresponds to the E-step in an EM algorithm. 

Taking the conditional expectation of the component negative log-likelihood and holding $\hat{y}_{rnk}$ fixed gives the surrogate loss
\begin{align}
 Q_k&=\sum_{r,n}\left[\frac{\hat{y}_{rnk}}{v_{rnk}} + \ln v_{rnk}\right]\\
&=\sum_{r,n}\left[\frac{\hat{y}_{rnk}\,e^{\lambda_k\tau_n}}{a_{rk}} + \ln a_{rk}-\lambda_k\tau_n\right].
 \label{eq:qfun}
\end{align}
For a candidate decay rate $\lambda$, define $M_{rk}(\lambda)=\sum_n\hat{y}_{rnk}\,e^{\lambda\tau_n}$ as the decay-compensated aggregate energy over all time frames. Although the surrogate is nonconvex in $a_{rk}>0$, its derivative changes sign only once, yielding the unique global amplitude minimum for fixed posterior powers and decay rate. Profiling out the amplitudes therefore gives
\begin{align}
 a_{rk}^{\star}(\lambda)&=\frac{M_{rk}(\lambda)}{N}, \label{eq:amplitude}\\ 
 \widetilde Q_k(\lambda)&=\sum_r\left[N\ln\frac{M_{rk}(\lambda)}{N}-\lambda B\right],\label{eq:profiled-surrogate}
\end{align}
where $B=\sum_n \tau_n$ and constants independent of $\lambda$ are omitted. The first and second derivatives of the profiled loss are
\begin{align}
 \widetilde Q_k'(\lambda)&=\sum_r\left[N\frac{M_{rk}'}{M_{rk}}-B\right],\notag\\
 \widetilde Q_k''(\lambda)&=\sum_rN\left[\frac{M_{rk}''}{M_{rk}}
 -\left(\frac{M_{rk}'}{M_{rk}}\right)^2\right]\geq0.
 \label{eq:curvature}
\end{align}
Differentiating $M_{rk}$ introduces factors $\tau_n$ and $\tau_n^2$ in its sum. The second derivative can be written as a sum of weighted time variance that is nonnegative.\footnote{Or one can prove this from Cauchy--Schwarz inequality.} For fixed posterior powers, the profiled surrogate is therefore convex in the scalar decay rate. We solve this subproblem using safeguarded Newton updates within positive bounds, then update the amplitudes using \eqref{eq:amplitude}. Optimizing $Q_k$ for decay rate and amplitudes corresponds to the M-step in EM algorithm.

One iteration sweeps over all components $k$ in turn, refreshing $\mathbf V$ from \eqref{eq:factorization} and posterior powers from \eqref{eq:estep} between updates. The optional noise update is $a_{r0}=N^{-1}\sum_n\hat{y}_{rn0}$. From positive initialization, accepting updates that decrease the component surrogate results in a non-increasing global loss over iterations~\cite{FesslerSpacealternatinggeneralized1994}. Global optimality of the joint fit is not guaranteed, but stationary points can be found~\cite{fevotteNonnegativeMatrixFactorization2009}.

\subsection{Contribution-weighted SAGE}

Because of different exponential decay rates, the components' power can vary in orders of magnitude. When a component contributes little in the total modeled variance $V_{rn}$, i.e. $\rho_{rnk}\ll1$, \eqref{eq:estep} gives a posterior power close to the current component variance. Such samples provide little information for updating the component parameters and can slow down the convergence speed. We introduce contribution based weighting factor in the surrogate from \eqref{eq:qfun} to emphasize more informative samples:
\begin{equation}
 w_{rnk}=\rho_{rnk}^{p},\quad
 Q_k^{(w)}=\sum_{r,n}w_{rnk}\left[\frac{\hat{y}_{rnk}}{v_{rnk}}+\ln v_{rnk}\right]. \label{eq:cw-qfun}
\end{equation}
For $p\geq0$, $p=0$ recovers standard SAGE; $p=1,2$ increasingly emphasize samples with large component contributions. We hold $w_{rnk}$ and $\hat{y}_{rnk}$ fixed during each update.

Define the weighted intermediate quantities
\begin{align*}
 M_{rk}^{(w)}&=\sum_n w_{rnk}\,\hat{y}_{rnk}\,e^{\lambda\tau_n},\\
 W_{rk}&=\sum_nw_{rnk},\qquad B_{rk}^{(w)}=\sum_nw_{rnk}\tau_n.\label{eq:cw-var}
\end{align*}
Replacing $(M_{rk},N,B)$ in \eqref{eq:amplitude}--\eqref{eq:curvature} by $(M_{rk}^{(w)},W_{rk},B_{rk}^{(w)})$ gives the amplitude and decay rate updates. Convexity in each decay rate update is retained. Refresh the contribution ratios between component updates. These weighted updates require positive $W_{rk}$; numerically negligible weights require a safeguard.

Reweighting the surrogate no longer guarantees descent or stationarity of the observed-data IS loss in \eqref{eq:loss}. In the reported experiments, however, CW-SAGE reduces the loss and RT estimation error faster than standard SAGE. The exponent $p$ controls the emphasis on strong component contributions, but imposes the deviation from the original IS fit. Subsequent refinement with $p=0$ is also possible.

\section{Results}
\label{sec:result}

\subsection{Experiments with synthetic data}

We first examine synthetic STFT coefficients generated from \eqref{eq:model} to compare standard SAGE ($p=0$) with CW-SAGE ($p=1,2$) to showcase the efficiency of our proposed algorithm. The representative example has true RTs of (0.80, 1.40)\,s. We choose $K=2,\,R=256$ and $N=256$. Assuming this is from a frequency bin of RIRs with sampling rate 48\,kHz and STFT hop size 256, the RIRs would be 1.37\,s long. For each RIR, the component amplitude $a_{r1}$ is sampled uniformly from $\mathcal U(0,1)$ and $a_{r2}=1-a_{r1}$, giving unit expected signal power at the time origin, which corresponds to 0\,dB. Noise-floor amplitudes are sampled from $\mathcal N(-40,1)$ in dB.

\begin{figure}[!htbp]
    \centering
  \includegraphics[width=\linewidth]{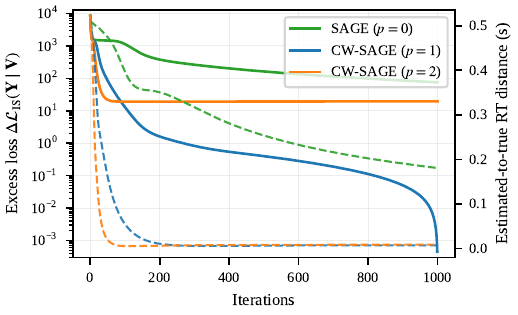}
  \caption{Excess IS loss (solid, left axis) and L2 RT-pair error (dashed, right axis) over SAGE iterations. Loss is shown relative to the lowest value attained by the displayed methods within 1000 iterations.}
  \label{fig:loss-convergence}
\end{figure}

\begin{figure}[!t]
  \centering
 \includegraphics[width=\linewidth]{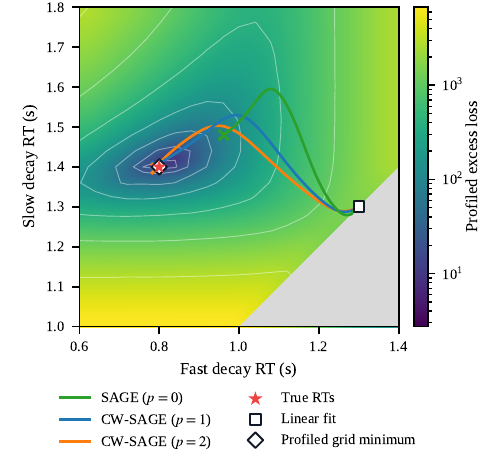}
 \caption{Amplitude-profiled observed-data IS loss and estimated RT trajectories from a pooled linear-fit initialization. The gray region is excluded due to parameter symmetry between the two RTs.}
 \label{fig:loss-surface}
\end{figure}

Figure~\ref{fig:loss-convergence} compares convergence in the observed-data IS objective and in the estimated RTs. We plot only the excess IS loss from the achieved best minimum $\Delta\mathcal{L}_{\text{IS}}=\mathcal{L}_{\text{IS}} - \min \mathcal{L}_{\text{IS}}$ for better visualization. CW-SAGE leaves the initial high-loss region substantially faster than standard SAGE, and its RT-pair error decreases correspondingly. The \(p=2\) variant approaches the true pair fastest initially, but \(p=1\) attains a lower final loss and slightly lower final RT error. Figure~\ref{fig:loss-surface} complements Fig.~\ref{fig:loss-convergence} and shows the CW-SAGE trajectories approach the true RT-pair with a smaller excursion. At each RT-pair, the plotted loss is minimized over amplitudes, profiling them out of parameter space.

To assess estimation accuracy across decay configurations, we repeat the experiment with additional RT-pairs. Across 100 randomly sampled two-decay configurations with $T_{1,2}\in[0.5,3]$\,s, the short- and long-RT RMSEs are 0.025 s and 0.047 s for $p=1$, and 0.039 s and 0.048 s for $p=2$. Recovery is less accurate when the true RTs are close, with a tendency to underestimate the shorter RT and overestimate the longer one.
For longer RTs, a given small error in decay rate produces a larger RT error through the nonlinear relationship in \eqref{eq:t60}. The largest RT-pair error occurs for the closely spaced pair (2.753, 2.783)\,s under $p=2$, where the L2 distance between estimated and true RT-pairs is 0.277\,s.

\subsection{Application to measured RIRs}

We next apply the model to measured RIRs from a coupled room transition dataset \cite{mckenzieAcousticAnalysisDataset2021}. The recordings we used are from a meeting room-to-hallway scene, with four sound source configurations: sources in the meeting room or hallway, with or without line of sight (LOS) from the sound source to all recording positions. Each configuration contains 101 recordings along a 5\,m receiver path crossing the doorway.\footnote{See \url{https://zenodo.org/records/6375520} for a detailed description of the dataset. v1.3 of the published data is used.} We assume two shared decay rates per frequency bin. The omnidirectional channel is sampled at 48\,kHz and analyzed using a 512-sample STFT window, a 256-sample hop, and a 512-point FFT. The first 10 STFT frames are discarded to select the late-reverberation segment. A stationary noise floor is included in the fit. The estimation is performed from CW-SAGE with $p=1$.

\begin{figure}[!tbp]
 \centering
 \includegraphics[width=\columnwidth]{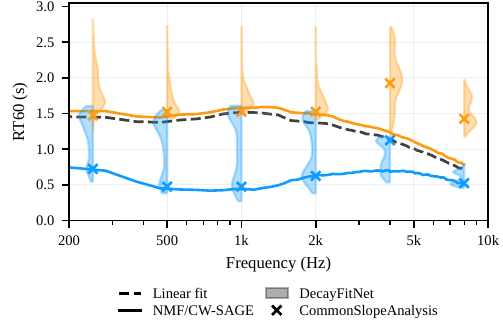}
 \caption{Estimated RTs for the room-to-hallway data. Blue and orange distinguish shorter and longer RTs.}
 \label{fig:real-t60}
\end{figure}

Figure~\ref{fig:real-t60} shows two distinct, smooth RT curves. The pooled linear fit used for initialization is included as a reference, alongside DecayFitNet and CommonSlopeAnalysis results~\cite{gotzNeuralNetworkMultiexponential2022,gotzCommonSlopeModelingLate2023}. At high frequencies, DecayFitNet produces a broad distribution of longer-RT estimates up to 2.5\,s, and clustered common-slope estimates differ from the jointly fitted curve. This comparison does not establish their relative accuracy without a ground-truth reference. However, the smooth frequency dependence obtained without cross-frequency regularization supports the practical value of joint estimation.


\begin{figure}
\centering\includegraphics[width=\columnwidth]{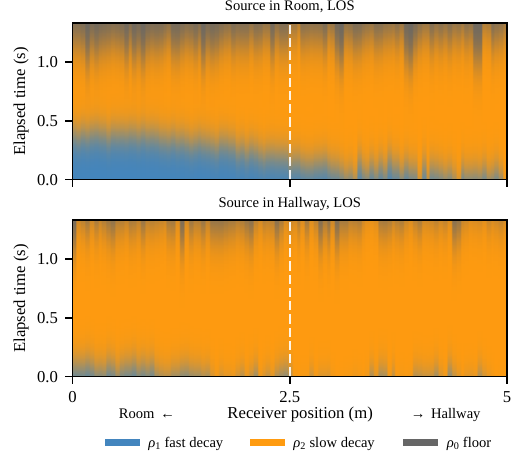}
    \caption{Space--time component contributions near 2\,kHz for the two configurations. The blending colors show the distribution of modeled variance shares between the decay components and noise floor. The dashed line marks the doorway between the room and hallway.}
    \label{fig:contribution}
\end{figure}

To illustrate the complementary support available to joint estimation, Fig.~\ref{fig:contribution} shows the component contribution ratios near a 2\,kHz bin. With the source in the room, the fast component contributes strongly at early times, particularly for receiver positions inside the room. With the source in the hallway, the slow component dominates most of the displayed space--time region. This pattern is consistent with stronger excitation of the shorter, room-associated decay when the source is inside the room, and of the longer, hallway-associated decay when the source is in the hallway. As the fast component attenuates, the slower component becomes increasingly prominent until the noise floor becomes significant. These differences explain the benefit of joint estimation: positions with strong component contributions help pin down decay rates that are difficult to identify from other RIRs alone.\footnote{See \url{https://github.com/jeremybbq/multi_slope_NMF} for code and full experiment results.}


\section{Discussion}
\label{sec:discussion}

Parameterized IS-NMF jointly estimates shared decay rates and spatial amplitudes directly from RIR STFT powers. In the synthetic experiments, CW-SAGE achieves faster loss and RT-error reduction per iteration than standard SAGE and accurately recovers well-separated decays. The measured results yield smooth frequency-dependent RT curves and reveal complementary space--time contributions, demonstrating the information available for decay rates update in the common-slope joint estimation.

The model assumes a prescribed number of decays and independent complex Gaussian coefficients. Closely spaced or weak components remain difficult to identify, while modal structure and correlations can limit the statistical approximation. Frequency-dependent modeling could further account for modal behavior at low frequencies and limited component identifiability at high frequencies.

Future work could establish a theoretical basis for CW-SAGE, including its convergence behavior and the choice of weighting exponent. A Bayesian extension could incorporate prior structure across frequency and space, quantify parameter uncertainty, and infer the number of decay components.

\balance
\bibliographystyle{IEEEbib}
\bibliography{references}

\end{document}